\documentclass[runningheads]{llncs}
\usepackage{graphicx}
\usepackage[american]{babel}

\usepackage{mathtools} 
\usepackage{algorithm}
\usepackage{algorithmic}
\usepackage{amsmath,amsfonts,amssymb}
\usepackage{url}
\usepackage{float}
\begin{document}
\title{PFGE: Parsimonious Fast Geometric Ensembling of DNNs
}
%
%
\author{Hao Guo\and
Jiyong Jin \and
Bin Liu\thanks{Correspondence author. This paper has been accepted by 19th Inter. Conf. on Intelligent Computing (ICIC 2023).}}
\authorrunning{H. Guo et al.}
%
\institute{Research Center for Applied Mathematics and Machine Intelligence,\\
Zhejiang Lab, Hangzhou, 311121 China \\
\email{\{guoh, jinjy, liubin\}@zhejianglab.com}}
\maketitle              
\begin{abstract}
Ensemble methods are commonly used to enhance the generalization performance of machine learning models. However, they present a challenge in deep learning systems due to the high computational overhead required to train an ensemble of deep neural networks (DNNs). Recent advancements such as fast geometric en-sembling (FGE) and snapshot ensembles have addressed this issue by training model ensembles in the same time as a single model. Nonetheless, these techniques still require additional memory for test-time inference compared to single-model-based methods. In this paper, we propose a new method called parsimonious FGE (PFGE), which employs a lightweight ensemble of higher-performing DNNs generated through successive stochastic weight averaging procedures. Our experimental results on CIFAR-$\{10,100\}$ and ImageNet datasets across var-ious modern DNN architectures demonstrate that PFGE achieves 5x memory efficiency compared to previous methods, without compromising on generalization performance. For those interested, our code is available at \url{https://github.com/ZJLAB-AMMI/PFGE}.
\end{abstract}
\section{Introduction}\label{sec:intro}
Ensemble methods are a popular way to enhance the generalization performance of machine learning models \cite{dietterich2000ensemble,zhou2012ensemble,caruana2004ensemble,dvzeroski2004combining}. However, their application with modern deep neural networks (DNNs) poses challenges. With millions or even billions of parameters, directly ensembling $k$ DNNs results in $k$-folded computational overhead in terms of both training time and memory requirements for test-time inference. Recently, fast geometric ensembling (FGE) and snapshot ensemble (SNE) methods have been proposed to overcome the hurdle of training time by enabling the training of DNN ensembles in the same time as a single model \cite{garipov2018loss,huang2017snapshot}. Nevertheless, these techniques still require higher memory overhead for test-time inference compared to single-model-based approaches. To reduce the test-time cost of ensembles, some researchers have proposed model compression and knowledge distillation methods \cite{bucilua2006model,hinton2015distilling}, which aim to train one single model that embodies the ``knowledge" of the ensembles. However, they do not account for the computational overhead of ensemble training.

This paper addresses the important challenge of reducing both training time cost and test-time memory budget for deep neural network (DNN) ensembling. This issue is especially critical for DNN applications with limited memory resources, such as edge computing on devices with restricted memory space \cite{zhou2021brief,varghese2016challenges,li2021model,zhang2023Commonsense}. We introduce a novel algorithm called PFGE that achieves 5x memory efficiency compared to prior ensemble methods for test-time inference, without sacrificing generalization performance or training efficiency. In other words, PFGE enables the reduction of both the training-time and test-time computational budgets for DNN en-sembling while maintaining high generalization performance.

The remainder of this paper is structured as follows. Section \ref{sec:related} presents related works. Section \ref{sec:PFGE} introduces the proposed PFGE algorithm. Section \ref{sec:experiments} presents experimental results. Finally, we conclude the paper in Section \ref{sec:conclusion}.
\section{Related Works}\label{sec:related}
Ensemble methods have traditionally posed significant computational challenges for learning with modern DNNs, due to the high overhead of both ensemble training and test-time inference. Nonetheless, researchers have made notable strides in adapting ensemble methods to DNNs, including FGE \cite{garipov2018loss}, SNE \cite{huang2017snapshot}, SWA-Gaussian (SWAG) \cite{maddox2019simple}, Monte-Carlo dropout \cite{gal2016dropout}, and deep ensembles \cite{fort2019deep,lakshminarayanan2017simple}. Of these, FGE, SNE, and SWAG are most relevant to this work, as they all employ a cyclical learning rate and enable the training of DNN ensembles in the same time as for one DNN model.

Both FGE and SNE construct DNN ensembles by sampling network weights from an SGD trajectory that corresponds to a learning rate \cite{smith2017cyclical}. Running SGD with a cyclical learning rate is equivalent in principle to performing SGD sampling with periodic warm restarts \cite{loshchilov2017sgdr}. Researchers have shown that the cyclical learning rate provides an efficient approach for collecting high-quality DNN weights that define the models for ensembling, as demonstrated by the authors of \cite{huang2017snapshot,garipov2018loss}.

Compared to SNE, FGE has a unique feature - a geometric explanation for its en-semble generation approach. Specifically, FGE is based on a geometric insight into the DNN loss landscape, which suggests that simple curves exist connecting local optima, over which both training and test accuracy remain approximately constant. FGE leverages this insight to efficiently discover these high-accuracy pathways between local optima.

Building upon the success of FGE, researchers proposed SWA, which averages high-performing network weights obtained by FGE for test-time inference \cite{izmailov2018averaging}. The underlying geometric insight of SWA is that averaging weights sampled from an SGD trajectory corresponding to a cyclical or constant learning rate can lead to wider optima in the DNN loss landscape, which results in better generalization \cite{keskar2016large}.
SWAG, on the other hand, uses the SWA solution as the center of a Gaussian distribu-tion to approximate the posterior of the network weights \cite{maddox2019simple}. The resulting model ensembles are generated by sampling from this Gaussian.
\section{The Proposed PFGE Algorithm}\label{sec:PFGE}
The design of PFGE is inspired by the observation that running a single stochastic weight averaging (SWA) procedure can lead to a higher-performing local optimum \cite{izmailov2018averaging}, and that running a series of SWA procedures successively may reveal a set of higher-performing weights than those obtained with SGD \cite{guo2022stochastic}. While FGE employs an ensemble of models found by SGD, we conjecture that using an ensemble of higher-performing models discovered by SWA, PFGE could achieve comparable generalization performance with far fewer models. As detailed in Section \ref{sec:experiments}, our experimental results indeed support this conjecture.

\begin{figure}[htbp]
\vskip 0.2in
\begin{center}
\centerline{\includegraphics[width=0.6\columnwidth]{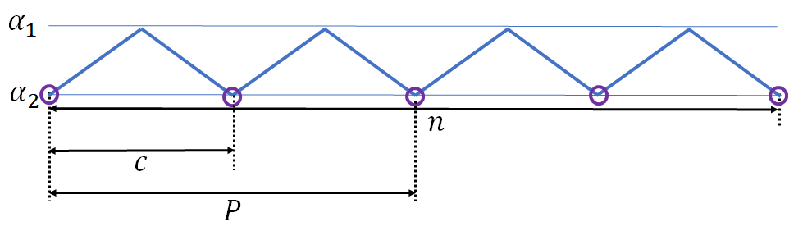}}
\caption{A conceptual diagram of the cyclical LR used by SWA, FGE and PFGE. The circles mask the time instances for recording the local optima discovered along the SGD trajectory. The real relationship between $c$, $P$, and $n$ is that $P$ is an integer multiple of $c$, and $n$ is an integer multiple of $P$. Here only one example case is plotted, in which $P=2c$ and $n=2P$. See the text for detailed explanations for all parameters involved.}
\label{fig:lrs}
\end{center}
\vskip -0.2in
\end{figure}

We provide the pseudo-codes for implementing SWA, FGE, and PFGE in Algorithms \ref{alg:swa}, \ref{alg:fge}, and \ref{alg:pfge}, respectively.
Each algorithm iteratively performs stochastic gradient-based weight updating, starting from a local optimum $w_0$ found by preceding SGD. The iterative weight updating operation uses a cyclical learning rate to allow the weight trajectory to escape from the current optimum, dis-cover new local optima, and converge to them. A graphical representation of the cyclical learning rate is shown in Figure \ref{fig:lrs}, where $\alpha_1$ and $\alpha_2$ define the minimum and maximum learning rate values, respectively, $c$ represents the cycle length, $n$ corresponds to the total number of allowable iterations that depends on the training time budget, and $P$ denotes the period for collecting member models for PFGE.
\begin{algorithm}[!htb]
\caption{SWA based model training and test-time prediction}
\label{alg:swa}
\textbf{Input}: initial network weights $w_{0}$, cyclical LR schedule $SC$, cycle length $c$, budget (the total number of allowable iterations) $n$, test data $x$\\
\textbf{Output}: predicted label $y$ of $x$
\begin{algorithmic}[1] 
\STATE $w\leftarrow w_{0}$; solution set $\mathcal{S}\leftarrow \{\}$.
\STATE $w_{\tiny{\mbox{SWA}}}\leftarrow w$.
\FOR{$i\leftarrow 1,2,\ldots,n$}
\STATE Compute current learning rate $\alpha$ according to $SC$.
\STATE $w\leftarrow w-\alpha\bigtriangledown \mathcal{L}_i(w)$ (stochastic gradient update).
\IF {mod($i$,$c$)=0}
\STATE $n_{\tiny{\mbox{models}}}\leftarrow i/c$ (number of models averaged).
\STATE $w_{\tiny{\mbox{SWA}}}\leftarrow \left(w_{\tiny{\mbox{SWA}}}\cdot n_{\tiny{\mbox{models}}}+w\right)/\left(n_{\tiny{\mbox{models}}}+1\right)$.
\ENDIF
\ENDFOR
\STATE Input $x$ into the DNN with weights $w_{\tiny{\mbox{SWA}}}$, then compute the its softmax output.
\STATE \textbf{return} $y$ that maximizes the above softmax output.
\end{algorithmic}
\end{algorithm}
\begin{algorithm}[!htb]
\caption{FGE based model training and test-time prediction}
\label{alg:fge}
\textbf{Input}: initial network weights $w_{0}$, cyclical LR schedule $SC$, cycle length $c$, budget (the total number of allowable iterations) $n$, test data $x$\\
\textbf{Output}: predicted label $y$ of $x$
\begin{algorithmic}[1] 
\STATE $w\leftarrow w_{0}$; solution set $\mathcal{S}\leftarrow \{\}$.
\FOR{$i\leftarrow 1,2,\ldots,n$}
\STATE Compute current learning rate $\alpha$ according to $SC$.
\STATE $w\leftarrow w-\alpha\bigtriangledown \mathcal{L}_i(w)$ (stochastic gradient update).
\IF {mod($i$,$c$)=0}
\STATE Add $w$ into $\mathcal{S}$ (collect weights).
\ENDIF
\ENDFOR
\STATE Given $x$ as the input, compute the average of softmax outputs of models included in $\mathcal{S}$.
\STATE \textbf{return} $y$ that maximizes the above averaged softmax output.
\end{algorithmic}
\end{algorithm}
\begin{algorithm}[!htb]
\caption{PFGE based model training and test-time prediction}
\label{alg:pfge}
\textbf{Input}: initial network weights $w_{0}$, cyclical LR schedule $SC$, cycle length $c$, budget (the total number of allowable iterations) $n$, test data $x$, model recording period $P$ \\
\textbf{Output}: predicted label $y$ of $x$
\begin{algorithmic}[1] 
\STATE $w\leftarrow w_{0}$; solution set $\mathcal{S}\leftarrow \{\}$.
\STATE $w_{\tiny{\mbox{SWA}}}\leftarrow w$.
\STATE $n_{\tiny{\mbox{recorded}}}\leftarrow 0$ (number of models recorded in $\mathcal{S}$).
\FOR{$i\leftarrow 1,2,\ldots,n$}
\STATE Compute current learning rate $\alpha$ according to $SC$.
\STATE $w\leftarrow w-\alpha\bigtriangledown \mathcal{L}_i(w)$ (stochastic gradient update).
\STATE $j\leftarrow i-n_{\tiny{\mbox{recorded}}}\times P$ (iterate index for the follow-up SWA procedure).
\IF {mod($j$,$c$)=0}
\STATE $n_{\tiny{\mbox{models}}}\leftarrow j/c$ (number of models that have been averaged within the current SWA procedure).
\STATE $w_{\tiny{\mbox{SWA}}}\leftarrow \left(w_{\tiny{\mbox{SWA}}}\cdot n_{\tiny{\mbox{models}}}+w\right)/\left(n_{\tiny{\mbox{models}}}+1\right)$.
\ENDIF
\IF {mod($i$,$P$)=0}
\STATE Add $w_{\tiny{\mbox{SWA}}}$ into $\mathcal{S}$ (collect weights).
\STATE $w\leftarrow w_{\tiny{\mbox{SWA}}}$ (initialization for the follow-up SWA procedure).
\STATE $n_{\tiny{\mbox{recorded}}}\leftarrow i/P$ (number of models recorded in $\mathcal{S}$).
\ENDIF
\ENDFOR
\STATE Given $x$ as the input, compute the average of softmax outputs of models recorded in $\mathcal{S}$.
\STATE \textbf{return} $y$ that maximizes the above averaged softmax output.
\end{algorithmic}
\end{algorithm}

The iterative weight updating operation employs a cyclical learning rate for letting the weight trajectory escape from the current optimum, discover new local optima, and converge to them. Figure \ref{fig:lrs} provides a graphical illustration of the cyclical learning rate, where $\alpha_1$ and $\alpha_2$ define the lower and upper bounds of the learning rate values, respectively, $c$ represents the cycle length, $n$ corresponds to the total number of allowable iterations that determine the training time budget, and $P$ denotes the period for storing member models used in PFGE.

As depicted in Algorithm \ref{alg:swa}, SWA maintains a running average of network weights collected at every $c$ iterations, and ultimately outputs a single model with the weight $w_{\tiny{\mbox{SWA}}}$ used for test-time inference. In practice, $w_{\tiny{\mbox{SWA}}}$ is derived as the average of $(\frac{n}{c})$ weights traversed along the SGD trajectory.

PFGE differentiates from FGE in the approach used to generate ensemble models. Unlike FGE, which employs purely SGD to generate its ensemble models (see operations 4 and 6 in Algorithm \ref{alg:fge}), PFGE uses multiple SWA operations performed successively (see operations 10 and 13 in Algorithm \ref{alg:pfge}) to generate the models. In the successively performed SWA procedures, the output of an SWA procedure is used as the initialization for its subsequent SWA operation (see operation 14 in Algorithm \ref{alg:pfge}). The code to implement PFGE is available at \url{https://github.com/ZJLAB-AMMI/PFGE}.
\section{Experiments}\label{sec:experiments}
We compare PFGE against prior state-of-the-art methods, including FGE \cite{garipov2018loss}, SWA \cite{izmailov2018averaging}, and SWAG \cite{maddox2019simple}, on widely-used image datasets such as CIFAR-100 \cite{krizhevsky2009learning} and ImageNet ILSVRC-2012 \cite{deng2009imagenet,russakovsky2015imagenet}, to evaluate its generalization performance for image classification. We also test its capability for uncertainty calibration, and conduct mode connectivity test for PFGE.
\subsection{Experimental Setting}\label{sec:experiment_setting}
As illustrated in Algorithms \ref{alg:swa}-\ref{alg:pfge}, SWA, FGE, and PFGE are all initialized with a local optimum $w_{0}$ and a learning rate schedule. For all architectures and datasets considered, we initial-ize all algorithms under comparison with the same $w_{0}$ and identical learning rate settings. Following \cite{garipov2018loss}, we employ a triangle learning rate schedule, as depicted in Figure \ref{fig:lrs}. Specifically, we set $c$ to iteration numbers corresponding to 2 or 4 epochs (following \cite{garipov2018loss}), $P$ to 10 epochs, and $n$ to 40 or 20 epochs. For $\alpha_1$, and $\alpha_2$, we adopt the same values as used in \cite{garipov2018loss}. We fix the mini-batch size for model training at 128.

For CIFAR-$\{10,100\}$, we obtain $w_{0}$ by running a standard SGD with momentum regulated by the same type of decaying learning rate schedule as employed in \cite{izmailov2018averaging}, until convergence to minimize an L2-regularized cross-entropy loss. We adopt the hyperparameters of SGD, such as the weight decay parameter and momentum factor, in the same manner as in \cite{izmailov2018averaging}. For ImageNet, we use pre-trained models ResNet-50 \cite{he2016deep}, ResNet-152 \cite{he2016deep}, and DenseNet-161 \cite{huang2017densely} available in PyTorch to initialize $w_{0}$, and set $n$ to be the iteration number corresponding to 40 epochs.

In our experiments, PFGE consistently employs four model components and uti-lizes the average of their softmax outputs for test-time prediction. Conversely, FGE and SWAG use the entire ensemble, consisting of 20 models, for test-time inference. To ensure that all algorithms under comparison have equal overhead in test-time inference, we compare PFGE against FGE$^{\star}$ and SWAG$^{\star}$, which are lightweight versions of FGE and SWAG, respectively. The sole distinction between FGE$^{\star}$ (resp. SWAG$^{\star}$) and FGE (resp. SWAG) is that the former generates test-time predictions based on the last four model components added to the ensemble set $\mathcal{S}$, while the latter utilizes all model components in $\mathcal{S}$. Our performance metric of interest is test accuracy.
\subsection{CIFAR Datasets}\label{sec:cifar}
We evaluate the performance of PFGE and competitor methods on various network architectures, including VGG16 \cite{simonyan2014very}, Preactivation ResNet-164 (PreResNet-164) \cite{he2016identity}, WideResNet-28-10 \cite{zagoruyko2016wide}, using datasets CIFAR-$\{10,100\}$.

We independently execute each algorithm at least three times and report the aver-age test accuracy results along with their corresponding standard errors in Tables \ref{table:big_table_cifar} and \ref{table:big_table_cifar100}. Our experimental findings indicate that on CIFAR-10, the VGG16 architecture achieves the highest test accuracy (93.41\%) using PFGE.
\begin{table*}
\centering
\caption{Test accuracy on CIFAR-10. We compare PFGE with FGE$^{\star}$, SWA, and SWAG$^{\star}$. Best results for each architecture are \textbf{bolded}. Results for FGE and SWAG, which use 5x memory resources for test-time inference compared with PFGE, are also listed for reference.}
\label{table:big_table_cifar}
\begin{tabular}{c|c} 
\hline
 & Test Accuracy (\%)  \\
\begin{tabular}{c} 
Algorithm \\ \hline PFGE   \\  FGE$^{\star}$   \\  SWA   \\ SWAG$^{\star}$ \\ \hline FGE \\ SWAG
\end{tabular} 
& \begin{tabular}{ccc} 
VGG16 & PreResNet & WideResNet \\ \hline
\textbf{93.41}$_{\pm0.08}$ & 95.70$_{\pm0.05}$ & 96.37$_{\pm0.03}$\\
93.03$_{\pm0.18}$ & 95.52$_{\pm0.08}$ & 96.14$_{\pm0.07}$\\
93.33$_{\pm0.02}$ & \textbf{95.78}$_{\pm0.07}$ & \textbf{96.47}$_{\pm0.04}$\\
93.24$_{\pm0.06}$ & 95.45$_{\pm0.14}$ & 96.36$_{\pm0.04}$\\ \hline
93.40$_{\pm0.08}$ & 95.57$_{\pm0.05}$ & 96.27$_{\pm0.02}$\\
93.37$_{\pm0.07}$ & 95.61$_{\pm0.11}$ & 96.45$_{\pm0.07}$\\
\end{tabular} 
\\ \hline
\end{tabular} 
\end{table*}
\begin{table*}
\centering
\caption{Test accuracy on CIFAR-100. We compare PFGE with FGE$^{\star}$, SWA, and SWAG$^{\star}$. Best results for each architecture are \textbf{bolded}. Results for FGE and SWAG, which use 5x memory resources for test-time inference compared with PFGE, are also listed for reference.}
\label{table:big_table_cifar100}
\begin{tabular}{c|c} 
\hline
 & Test Accuracy (\%) \\
\begin{tabular}{c} 
Algorithm \\ \hline PFGE   \\  FGE$^{\star}$   \\  SWA   \\ SWAG$^{\star}$ \\ \hline FGE \\ SWAG
\end{tabular} 
& \begin{tabular}{ccc} 
VGG16 & PreResNet & WideResNet \\ \hline
\textbf{74.17}$_{\pm0.04}$ & \textbf{80.06}$_{\pm0.13}$ & \textbf{81.96}$_{\pm0.01}$\\
73.49$_{\pm0.24}$ & 79.76$_{\pm0.06}$ & 81.09$_{\pm0.25}$\\
73.83$_{\pm0.20}$ & 79.97$_{\pm0.06}$ & 81.92$_{\pm0.02}$\\
73.77$_{\pm0.18}$ & 79.24$_{\pm0.04}$ & 81.55$_{\pm0.06}$\\ \hline
74.34$_{\pm0.05}$ & 80.17$_{\pm0.09}$ & 81.62$_{\pm0.16}$\\
74.15$_{\pm0.17}$ & 80.00$_{\pm0.03}$ & 81.83$_{\pm0.12}$\\
\end{tabular} 
\\ \hline
\end{tabular} 
\end{table*}

Regarding PreResNet-164 and WideResNet-28-10 architectures, our findings demonstrate that PFGE outperforms FGE$^{\star}$ and SWAG$^{\star}$, but performs slightly worse than SWA in terms of test accuracy; SWA achieves the highest accuracy (95.78\% and 96.47\%).
On CIFAR-100, we observe that PFGE delivers the best performance in terms of test accuracy for all network architectures. Furthermore, as evident from Tables \ref{table:big_table_cifar} and \ref{table:big_table_cifar100}, even when compared with FGE and SWAG, which utilize the full ensemble of model components for test-time inference, PFGE attains comparable or superior performance in terms of test accuracy while only requiring a 20\% memory overhead for test-time inference.
\begin{table*}
\centering
\caption{Test accuracy on Imagenet. We compare PFGE with FGE$^{\star}$, SWA, and SWAG$^{\star}$. Best results for each architecture are \textbf{bolded}. Results for FGE and SWAG, which use 5x memory resources for test-time inference compared with PFGE, are also listed for reference.}
\label{table:big_table_imagenet}
\begin{tabular}{c|c} 
\hline
 & Test Accuracy (\%)\\
\begin{tabular}{c} 
Algorithm \\ \hline PFGE   \\  FGE$^{\star}$   \\  SWA   \\ SWAG$^{\star}$ \\ \hline FGE \\ SWAG
\end{tabular} 
& \begin{tabular}{ccc} 
ResNet-50 & ResNet-152 & DenseNet-161 \\ \hline
\textbf{77.06}$_{\pm0.19}$ & \textbf{79.07}$_{\pm0.04}$ & \textbf{78.72}$_{\pm0.08}$\\
76.85$_{\pm0.07}$ & 78.73$_{\pm0.02}$ & 78.53$_{\pm0.08}$\\
76.70$_{\pm0.38}$ & 78.82$_{\pm0.02}$ & 78.41$_{\pm0.29}$\\
76.19$_{\pm0.29}$ & 78.72$_{\pm0.04}$ & 77.19$_{\pm0.83}$\\ \hline
77.17$_{\pm0.08}$ & 79.13$_{\pm0.06}$ & 78.91$_{\pm0.06}$\\
76.70$_{\pm0.35}$ & 79.10$_{\pm0.06}$ & 77.94$_{\pm0.62}$\\
\end{tabular} 
\\ \hline
\end{tabular} 
\end{table*}
\subsection{IMAGENET}\label{sec:imagenet}
We experiment with network architectures ResNet-50 \cite{he2016deep}, ResNet-152 \cite{he2016deep}, and DenseNet-161 \cite{huang2017densely} on ImageNet ILSVRC-2012 \cite{deng2009imagenet,russakovsky2015imagenet}. As in our previous experiments, we execute each algorithm three times independently.

The results are summarized in Table \ref{table:big_table_imagenet}, which reveals that PFGE surpasses FGE$^{\star}$ and SWAG$^{\star}$ in terms of test accuracy and achieves comparable performance to FGE and SWAG.
\subsection{Performance of separate models in the ensemble}
We evaluate the performance of each individual model in PFGE and FGE. Each sep-arate model can be viewed as a ``snapshot" of the SGD trajectory.

\begin{figure}[H]
\centering
\includegraphics[width=0.5\linewidth]{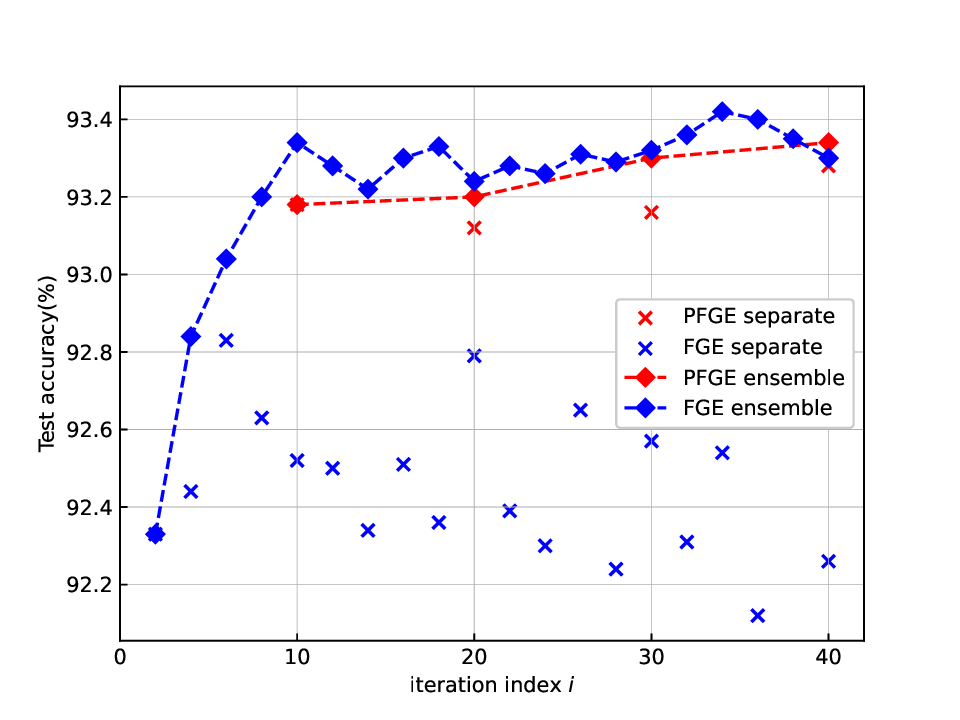}\includegraphics[width=0.5\linewidth]{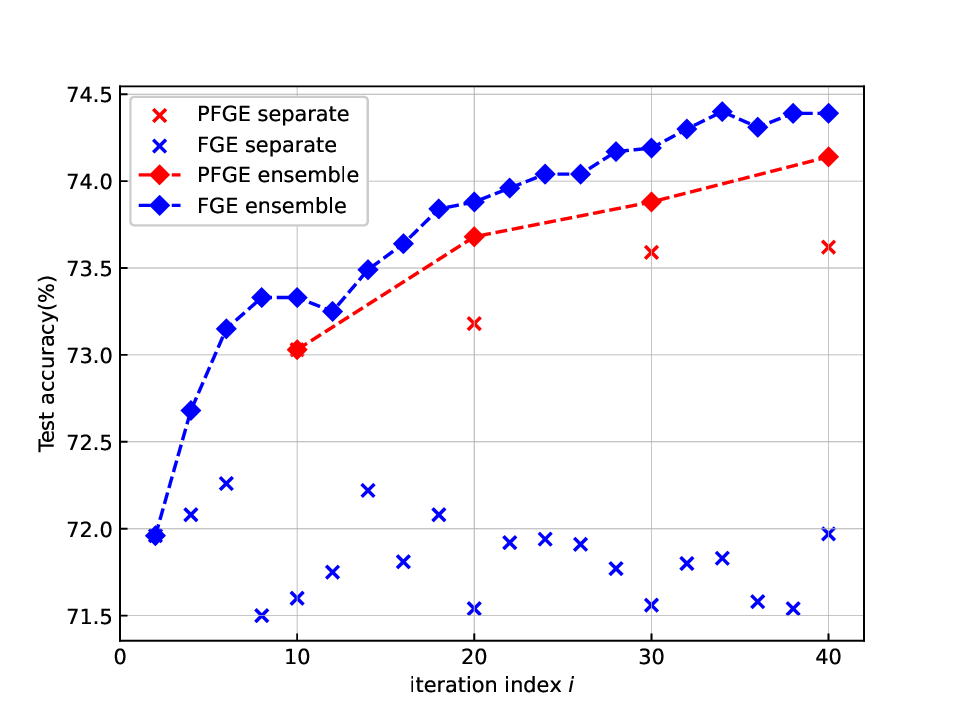}\\
\includegraphics[width=0.5\linewidth]{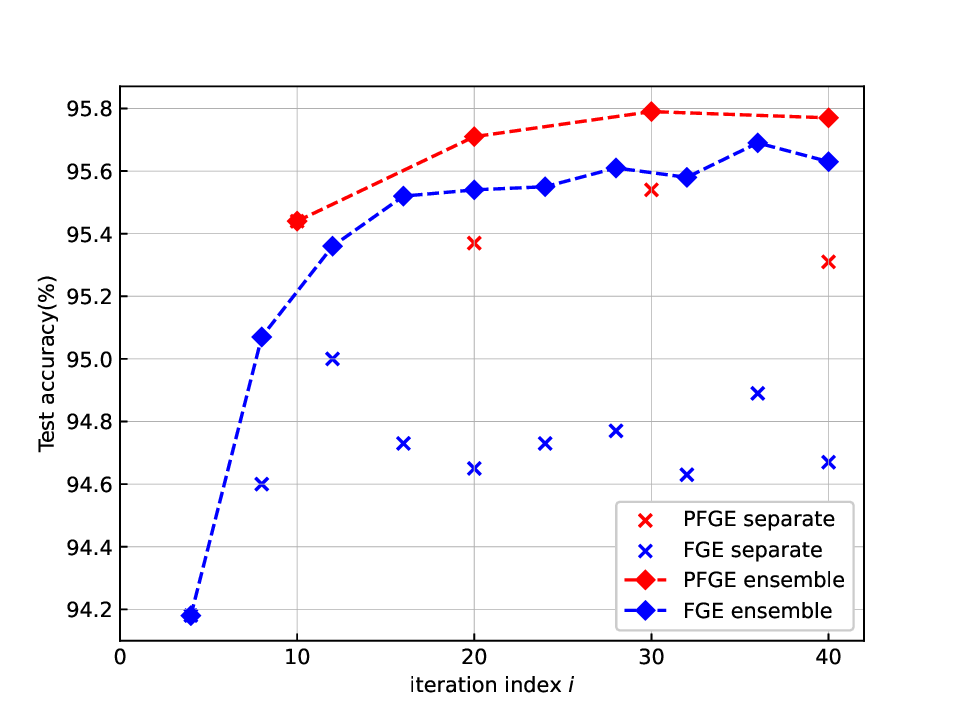}\includegraphics[width=0.5\linewidth]{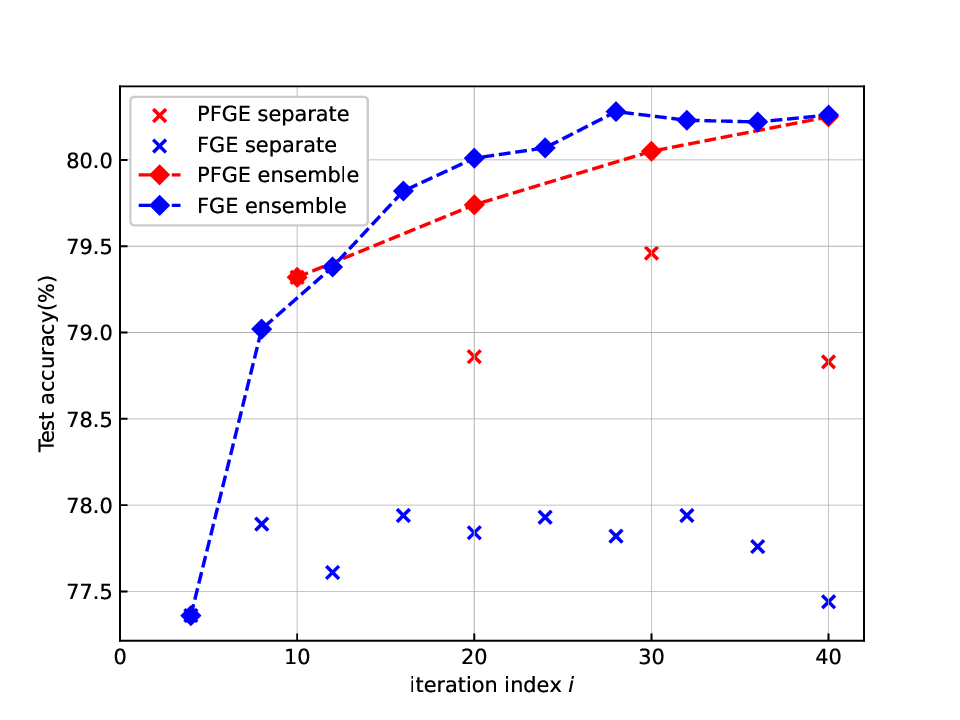}\\
\includegraphics[width=0.5\linewidth]{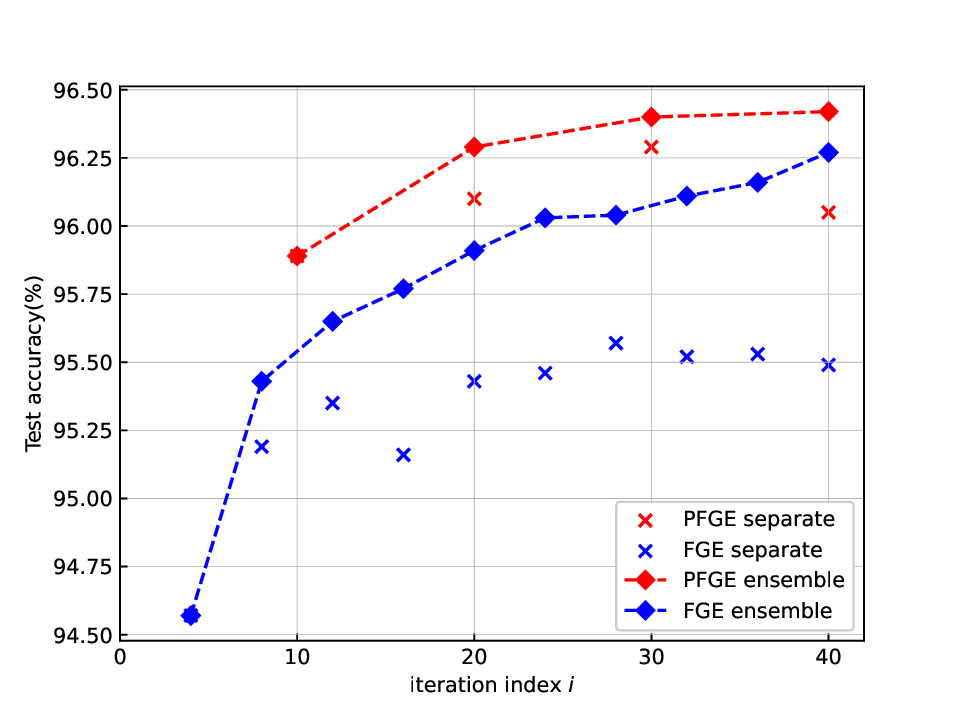}\includegraphics[width=0.5\linewidth]{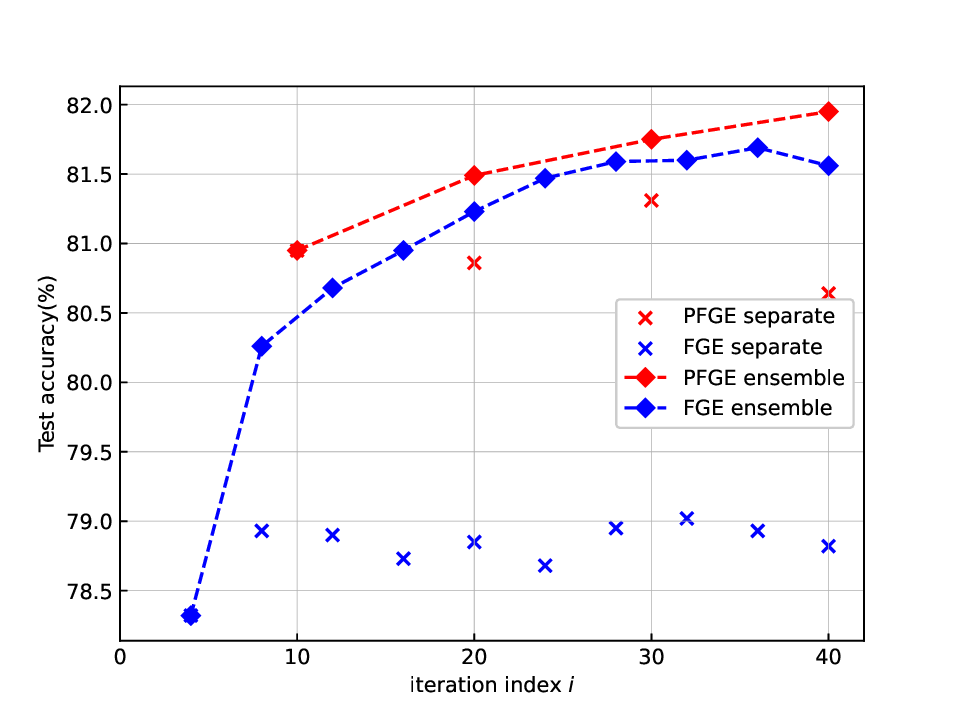}
\caption{Ensemble performance of PFGE and FGE as a function of the training iteration index $i$. We see that PFGE only uses 4 model components, while achieves a performance on par with or even better than that of FGE which uses 20 model components. Crosses represent the performance of separate ``snapshot" models, and diamonds show the performance of the ensembles composed of all models available at the given iteration. Left column: CIFAR-10. Right column: CIFAR-100. Top row: VGG16. Middle Row: PreResNet-164. Bottom row: WideResNet-28-10.}\label{fig:separate}
\end{figure}

To evaluate the generalization performance of these ``snapshot" models, we col-lect them at various training checkpoints and measure their test accuracy. The results are presented in Fig.\ref{fig:separate}, from which we observe that:
\begin{itemize}
  \item –	the individual models of PFGE outperform those of FGE in terms of test accuracy for all network architectures and both datasets considered;
  \item –	for both PFGE and FGE, the test accuracy of the model ensemble increases with an increase in the number of member models utilized.
\end{itemize}
Our experimental findings suggest that due to the superior quality of separate model components, PFGE outperforms FGE and SWAG when they are equipped with an equal number of model components.
\subsection{On training efficiency and test-time cost}
As highlighted in Section \ref{sec:experiment_setting}, all methods employ the same learning rate setting, execute an equal number of SGD iterations, and possess identical training efficiency (in terms of training time). Their test-time cost, primarily in terms of memory overhead, is proportional to the number of member models utilized, represented here by $K$. For PFGE, we have $K=(\frac{n}{P})=4$. In contrast, for FGE \cite{garipov2018loss} and SWAG \cite{maddox2019simple}, we have $K=(\frac{n}{c})=20$ according to our experimental setting in Section \ref{sec:experiment_setting}. This implies that using PFGE necessitates only a $\frac{4}{20}=20\%$ memory overhead for test-time inference, compared with FGE \cite{garipov2018loss} and SWAG \cite{maddox2019simple}. Concerning test-time efficiency, SWA \cite{izmailov2018averaging} is the preferred option since it always produces a single model ($K=1$) for use. However, in comparison to ensemble methods, SWA faces issues with uncertainty calibration \cite{maddox2019simple,guo2022optimization}, a crucial aspect closely linked to detecting out-of-distribution samples \cite{maddox2019simple,guo2022optimization}.
\subsection{Mode Connectivity Test}\label{sec:connectivity}
In a related study \cite{yang2021taxonomizing}, researchers illustrate that ensembles of trained models tend to converge to locally smooth regions of the loss landscape, resulting in the highest test accuracy. This indi-cates that the mode connectivity of model components closely influences the en-semble's generalization performance. To determine the mode connectivity of PFGE and FGE model components, we conducted an experiment and discovered that the mode connectivity of PFGE's model components is superior to that of FGE.

Specifically, we begin by randomly selecting two adjacent model components, $w$ and $w'$, from the ensemble set$\mathcal{S}$. We then search for a low-energy curve, denoted by $\gamma(t)$, where $t\in[0,1]$, that connects $w$ and $w'$ such that $\gamma(0)=w$ and $\gamma(1)=w'$. This curve is found by minimizing $\int_0^1\mathcal{L}(\gamma(t))dt$, where $\mathcal{L}$ represents the loss function of the DNN. As in \cite{yang2021taxonomizing,garipov2018loss}, we approximate this integral with $\mathbb{E}_{t\sim U(0,1)}\mathcal{L}(\gamma_{\phi}(t))$ using Bezier curve.  To construct the Bezier curve, which has $k+1$ bends, we use the formula $\gamma_{\phi}(t)=\sum_{j=0}^{k}\binom{k}{j}(1-t)^{k-j}t^jw_j$ for $t\in[0,1]$, $U(0,1)$ denotes a continuous uniform distribution between 0 and 1, and $w_0=w, w_k=w'$ are the endpoints of the curve. Additionally, $\phi$ consists of the parameters of additional models to be trained, namely $\{w_1,\ldots,w_{k-1}\}$. Finally, the mode connectivity between models $w$ and $w'$ is defined as per \cite{yang2021taxonomizing}
\begin{equation}\label{eqn:mc}
\mbox{mc}(w,w')=\frac{1}{2}(\mathcal{L}(w)+\mathcal{L}(w'))-\mathcal{L}(\gamma_{\phi}(t^{\star})),
\end{equation}
where $t^{\star}$ maximizes the function $f(t)\triangleq |\frac{1}{2}(\mathcal{L}(w)+\mathcal{L}(w'))-\mathcal{L}(\gamma_{\phi}(t))|$.

\begin{figure}[H]
\centering
\includegraphics[width=0.5\linewidth]{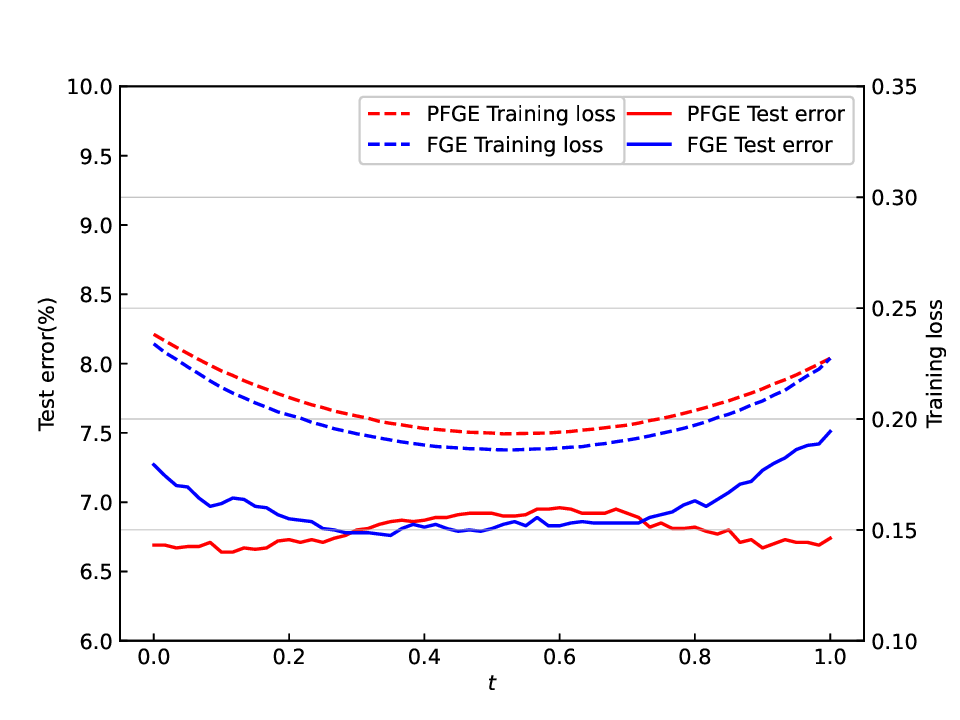}\includegraphics[width=0.5\linewidth]{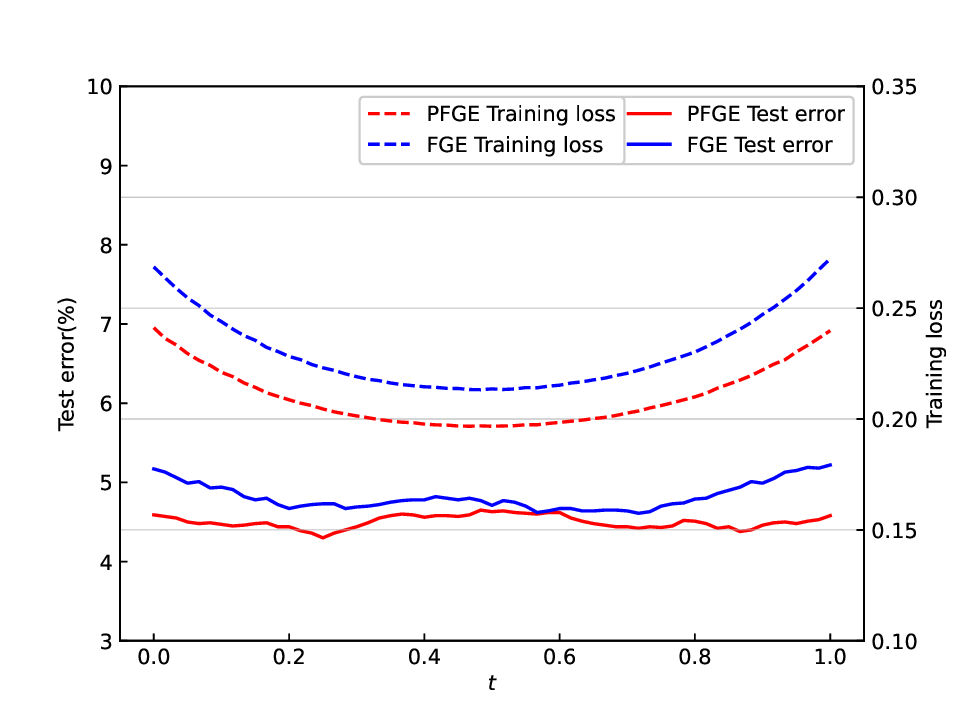}\\
\includegraphics[width=0.5\linewidth]{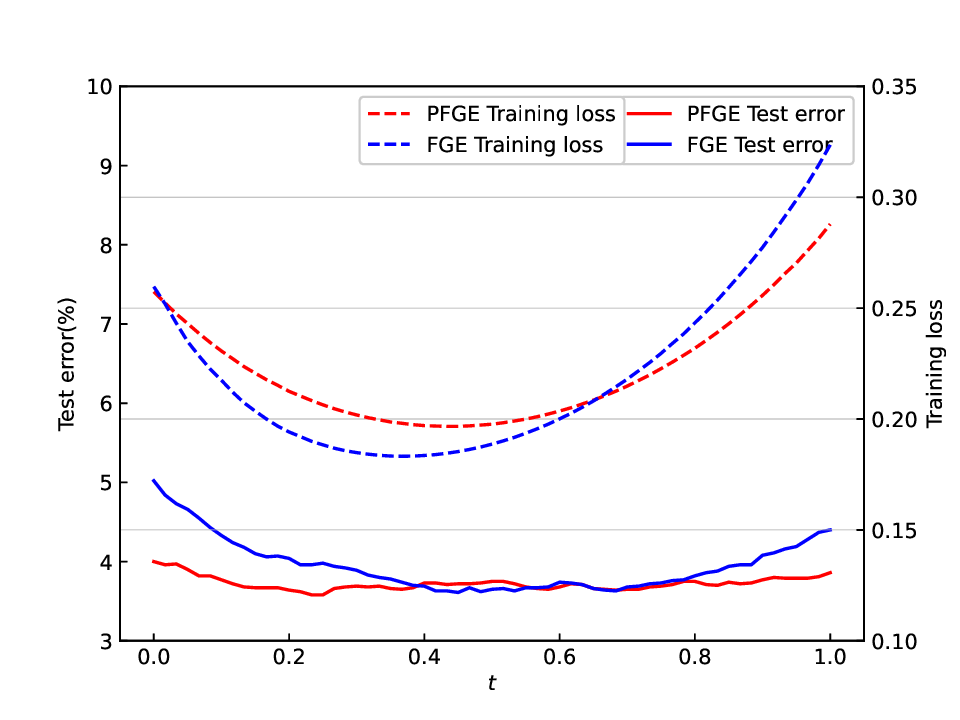}
\caption{Test error and training loss on CIFAR-10 corresponding to $\gamma_{\phi}(t)$ as a function of $t$. Top left: VGG16. Top right: PreResNet-164. Bottom: WideResNet-28-10.}\label{fig:mode_connect}
\end{figure}

We follow the same computational procedure and use the hyperparameter settings outlined in \cite{yang2021taxonomizing,garipov2018loss} to minimize the loss on the curve. We record the values of training loss and test error as functions of $t$, which we plot in Figure \ref{fig:mode_connect}. Tables \ref{table:stat_test} and \ref{table:stat_train} present the related statistics. Our findings, as seen in Figure \ref{fig:mode_connect}, indicate that PFGE has a consistently lower energy training loss curve than FGE for both VGG16 and PreResNet-164. Furthermore, for most $t$ values across all architectures, the test error curve of PFGE is below that of FGE. These results are also reflected in the consistent advantage of PFGE over FGE in terms of both test error and training loss, as evidenced by Tables \ref{table:stat_test} and \ref{table:stat_train}.
\begin{table}[H]
\centering
\caption{Statistics for test errors (\%) of $\gamma_{\phi}(t)$ on CIFAR-10. The test error values are collected along the changes of the $t$ value from 0 to 1.}
\label{table:stat_test}
\begin{tabular}{c|c|c|c} 
\hline
 & Max & Min & Mean\\
\begin{tabular}{c} 
Architecture \\ \hline VGG16 \\ PreResNet \\ WideResNet
\end{tabular} 
& \begin{tabular}{cc} 
PFGE & FGE \\ \hline
6.96 & 7.51 \\
4.65 & 5.22 \\
4.14 & 4.53
\end{tabular} 
& \begin{tabular}{cc} 
PFGE & FGE \\ \hline
6.64 & 6.76 \\
4.30 & 4.61 \\
3.43 & 3.63
\end{tabular} 
& \begin{tabular}{cc} 
PFGE & FGE \\ \hline
6.76 & 6.99 \\
4.50 & 4.82 \\
3.63 & 3.87
\end{tabular} 
\\ \hline
\end{tabular} 
\end{table}
\begin{table}[H]
\centering
\caption{Statistics for training losses of $\gamma_{\phi}(t)$ on CIFAR-10. The training loss values are collected along the changes of the $t$ value from 0 to 1.}
\label{table:stat_train}
\begin{tabular}{c|c|c|c} 
\hline
 & Max & Min & Mean\\
\begin{tabular}{c} 
Architecture \\ \hline VGG16 \\ PreResNet \\ WideResNet
\end{tabular} 
& \begin{tabular}{cc} 
PFGE & FGE \\ \hline
0.238 & 0.234 \\
0.241 & 0.272 \\
0.284 & 0.324
\end{tabular} 
& \begin{tabular}{cc} 
PFGE & FGE \\ \hline
0.193 & 0.186 \\
0.197 & 0.213 \\
0.217 & 0.238
\end{tabular} 
& \begin{tabular}{cc} 
PFGE & FGE \\ \hline
0.207 & 0.199 \\
0.210 & 0.230 \\
0.238 & 0.266
\end{tabular} 
\\ \hline
\end{tabular} 
\end{table}
\begin{table}[H] 
\centering
\caption{Mode connectivity test on CIFAR-10. See the definition of $\mbox{mc}$ in Equation (\ref{eqn:mc}). A $\mbox{mc}$ value closer to 0 indicates a better mode connectivity and vice versa. See more details on the relationship between the value of $\mbox{mc}$ and mode connectivity in \cite{yang2021taxonomizing}. The best results for each architecture are \textbf{bolded}.}
\label{table:mode_connect}
\begin{tabular}{c|c} 
\hline
 & $\mbox{mc}$ value \\
\begin{tabular}{c} 
Algorithm \\ \hline PFGE   \\  FGE
\end{tabular} 
& \begin{tabular}{ccc} 
VGG16 & PreResNet & WideResNet \\ \hline
\textbf{0.039} & \textbf{0.044} & \textbf{0.065}\\
0.045 & 0.057 & 0.084\\
\end{tabular} 
\\ \hline
\end{tabular} 
\end{table}

\begin{figure}[H]
\centering
\includegraphics[width=0.5\linewidth]{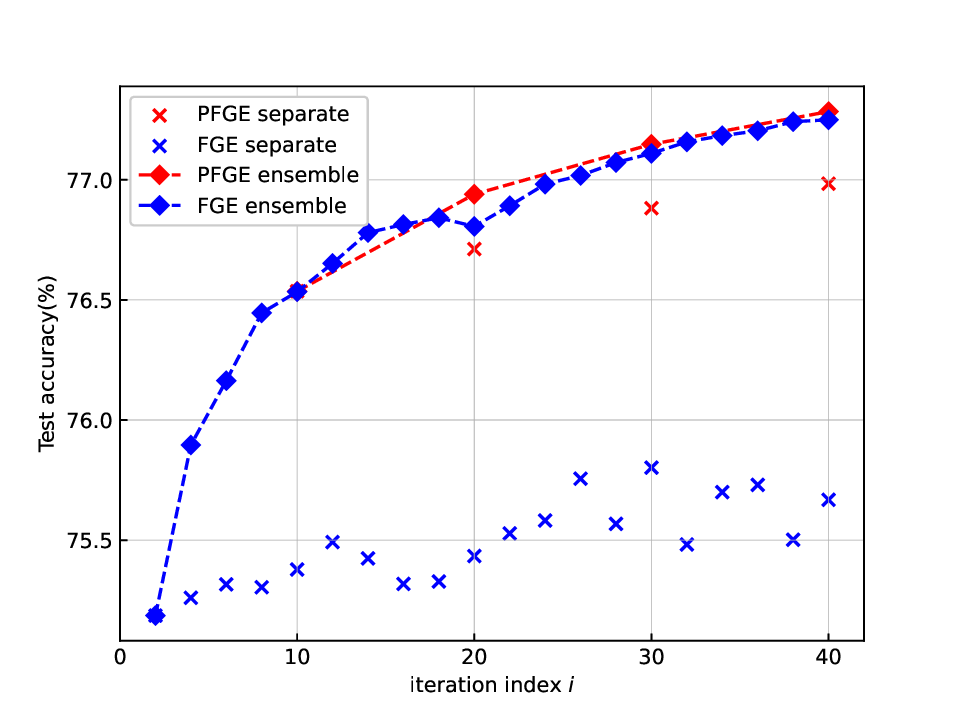}\includegraphics[width=0.5\linewidth]{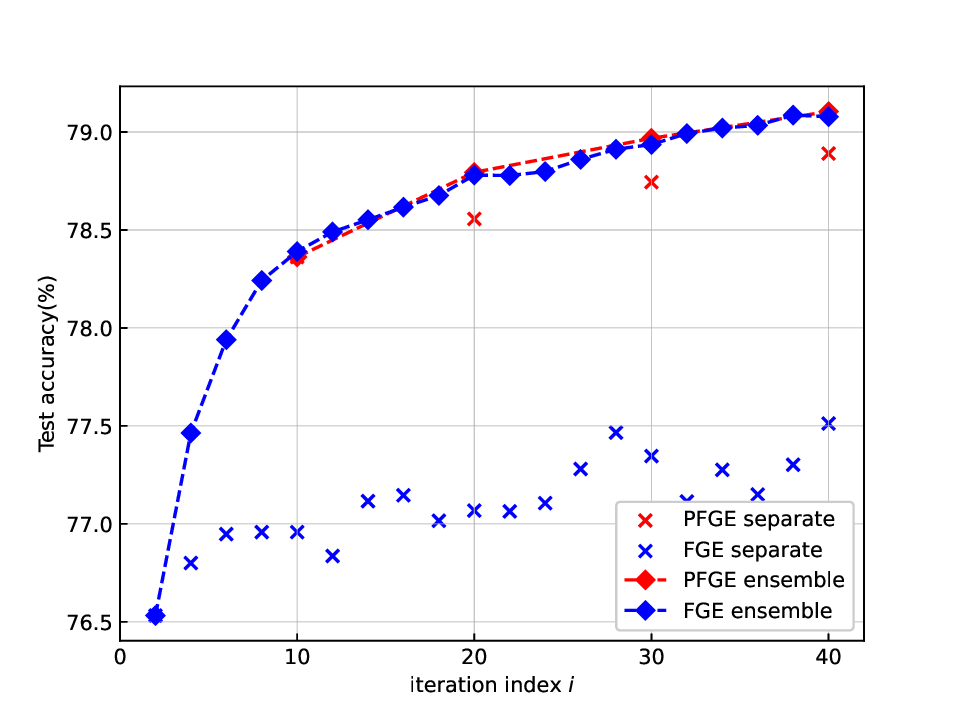}\\
\includegraphics[width=0.5\linewidth]{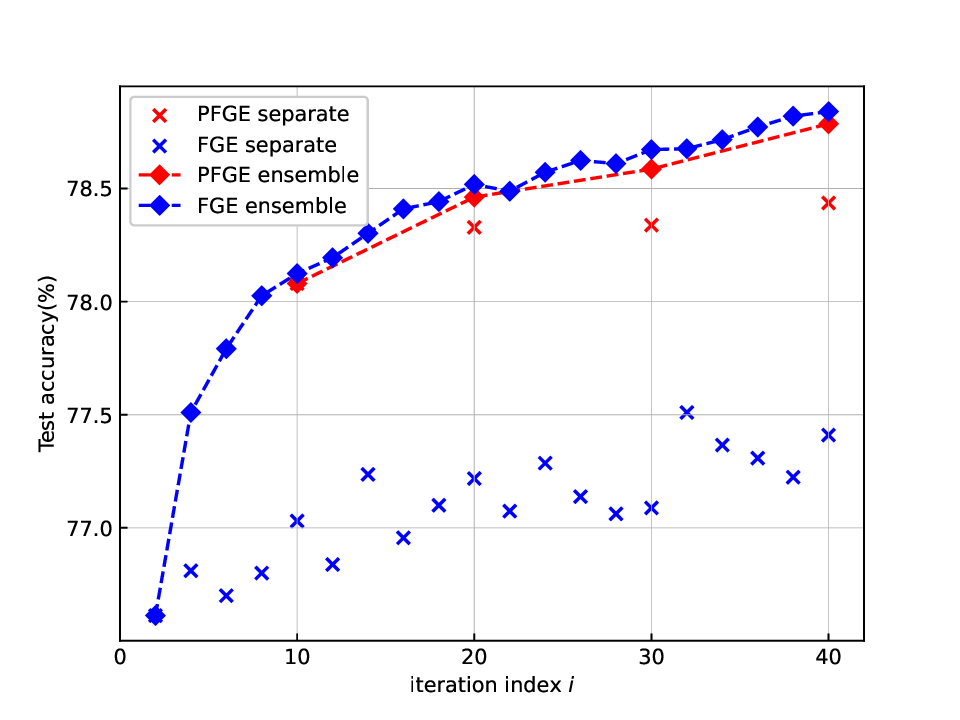}
\caption{Ensemble performance of PFGE and FGE on Imagenet as a function of the training iteration index $i$. We see that PFGE only uses 4 model components, while achieves a performance on par with that of FGE which uses 20 model components. Top Left: ResNet-50. Top Right: ResNet-152. Bottom: DenseNet-161. Crosses represent the performance of separate ``snapshot" models, and diamonds show the performance of the ensembles composed of all models available at the given iteration.}\label{fig:res_imagenet}
\end{figure}

The corresponding mc values, which are presented in Table \ref{table:mode_connect}, suggest that the model connectivity of PFGE exceeds that of FGE across all architectures.
\section{Conclusions}\label{sec:conclusion}
Ensemble methods are widely applied to enhance the generalization performance of machine learning models. However, their use with modern DNNs presents challenges due to the extensive computational overhead required to train the DNN ensemble. Recent advancements such as FGE and SWAG have addressed this issue by training model ensembles in a timeframe equivalent to that required for a single model. Nev-ertheless, these techniques still require extra memory for test-time inference when compared to single-model-based approaches. This paper introduces a new method called parsimonious FGE (PFGE), which is based on FGE but utilizes a lightweight ensemble of higher-performing DNNs produced via successive SWA procedures. Our experimental findings on CIFAR-10, CIFAR-100, and ImageNet datasets across various state-of-the-art DNN architectures demonstrate that PFGE can achieve up to five times memory efficiency compared to prior art methods such as FGE and SWAG without sacrificing generalization performance. We tested the mode connectivity of model components given by PFGE, finding that it is better than that of FGE. This provides a possible geometric explanation for why PFGE beats FGE if they are equipped with an equal number of model components.
\section*{Acknowledgment}
This work was supported by Research Initiation Project (No.2021KB0PI01) and Ex-ploratory Research Project (No.2022RC0AN02) of Zhejiang Lab.
\bibliographystyle{splncs04}
\bibliography{mybibfile}
\end{document}